\documentclass[letterpaper, 10 pt, conference]{ieeeconf}  % Comment this line out if you need a4paper
\AtBeginDocument{\let\autocite\cite}

\IEEEoverridecommandlockouts                              % This command is only needed if                                                           % you want to use the \thanks command

\usepackage{balance}
\usepackage{url}
\usepackage{cite}
\usepackage{graphicx} % for pdf, bitmapped graphics files
\usepackage{flushend}
\usepackage{booktabs}
\usepackage{tikz}

\newcommand\copyrighttext{%
  \footnotesize \textcopyright 2012 IEEE. Personal use of this material is permitted.
  Permission from IEEE must be obtained for all other uses, in any current or future
  media, including reprinting/republishing this material for advertising or promotional
  purposes, creating new collective works, for resale or redistribution to servers or
  lists, or reuse of any copyrighted component of this work in other works.}
\newcommand\copyrightnotice{%
\begin{tikzpicture}[remember picture,overlay]
\node[anchor=south,yshift=10pt] at (current page.south) {\fbox{\parbox{\dimexpr\textwidth-\fboxsep-\fboxrule\relax}{\copyrighttext}}};
\end{tikzpicture}%
}

\begin{document}
\IEEEoverridecommandlockouts
\overrideIEEEmargins

\title{\LARGE \bf
  Tactile Comfort: Lowering Heart Rate Through Interactions \\with a Pocket Robot
}

\author{Morten Roed Frederiksen$^{1}$, Kasper Stoy$^{2}$, and Maja Matari\'{c}$^{3}$
  \thanks{{$^{1}$Morten Roed Frederiksen {\tt\small mrof@itu.dk} and $^{3}$Maja Matari\'{c} are affiliated with the Interaction Lab at the University of Southern California. $^{2}$Kasper Stoy \tt\small ksty@itu.dk} is affiliated with the Computer Science Department of The IT-University of Copenhagen.}
}

\maketitle
\copyrightnotice
\begin{abstract}
Children diagnosed with anxiety disorders are taught a range of strategies to navigate situations of heightened anxiety. Techniques such as deep breathing and repetition of mantras are commonly employed, as they are known to be calming and reduce elevated heart rates.
Although these strategies are often effective, their successful application relies on prior training of the children for successful use when faced with challenging situations. This paper investigates a pocket-sized companion robot designed to offer a relaxation technique requiring no prior training, with a focus on immediate impact on the user's heart rate. The robot utilizes a tactile game to divert the user's attention, thereby promoting relaxation. We conducted two studies with children who were not diagnosed with anxiety:  a 14-day pilot study with two children (age 8) and a main study with 18 children (ages 7-8). Both studies employed a within-subjects design and focused on measuring heart rate during tactile interaction with the robot and during non-use. Interacting with the robot was found to significantly lower the study participants' heart rate (p$<$0.01) compared to the non-use condition, indicating a consistent calming effect across all participants. These results suggest that tactile companion robots have the potential to enhance the therapeutic value of relaxation techniques.
\end{abstract}

\section{Introduction}

In recent years, the integration of robotics into therapeutic settings has emerged as a frontier in healthcare innovation, blending technology with human care to address both psychological and physiological needs of users \cite{Rice2023TheEO,Bharatharaj2017SociopsychologicalAP,chang2013robot, Brivio2021ASM}. Examples of scenarios where robots impact the psychological well being of humans range from stress relief training robots \cite{Zhu2020EffectOS}, to soft therapeutic robots \cite{Hayashi2019ImportanceOS}, to focus-supporting college study partners for students with symptoms of attention deficit hyperactivity disorder \cite{oconnelmataric2024}.
Complex settings such as therapy sessions have seen robots introduced as therapy guides with promising results \cite{dino2019delivering}. Also Large Language Models have been shown to increase the adherence to daily practise routines as a part of Cognitive behavioral therapy (CBT) \cite{Kian2024CanAL} which is the current standard clinically proven treatment for children suffering from a anxiety disorder \cite{Tse2023SchoolbasedCT}. However, viable alternatives outside of the coping strategies, introduced in CBT to manage high anxiety situations, are still relatively limited \cite{avants1990, CartwrightHatton2004SystematicRO}. Furthermore, studies have shown that some patients still suffer from anxiety after having been exposed to CBT \cite{Lau2016ChildhoodAD}. 
This paper investigates the effectiveness of a pocket-sized companion robot in reducing a child's heart rate by conducting a pilot study and then a full study with children aged 7 to 8. 

\begin{figure}[h]
\centering \includegraphics[width=0.50\textwidth]{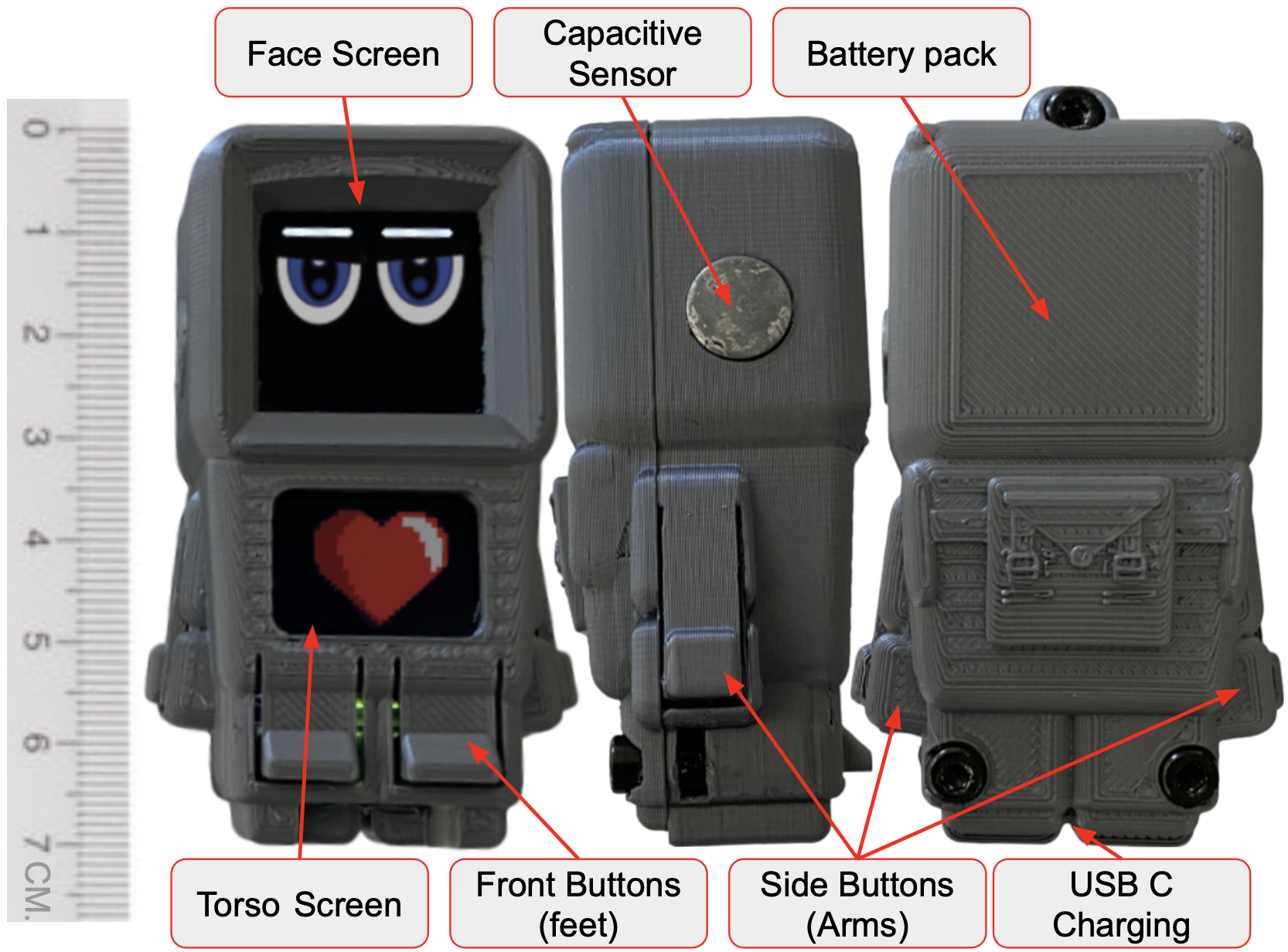} \caption{The "AffectaPocket" therapeutic pocket-sized robot developed to aid children suffering from an anxiety disorder. AffectaPocket is controlled by an onboard EPS32 chip, and is powered by a 3.7v battery. Children interact with the robot through a tactile rhythm game played by grasping the robot.}
\label{affecta_annotated}
\end{figure}

%\subsection*{Robot design}
We designed and built a pocket-sized robot, "AffectaPocket", that plays a simple tactile game with the user (see Figure \ref{affecta_annotated}). The robot has a 3D-printed exterior and four buttons (one on each side, two on the front) to record tactile input from the user's grasping hand.
The users initiates a tactile game by grasping the robot and in response, the robot generates a three-note pattern, and communicates it silently to the user through tactile vibration. The game consists of the user replicating the rhythmic pattern by pressing the robot's side buttons so that the timing of the presses matches the generated sequence. 

%\subsection*{Evaluating the physical impact}
%We evaluated the robot's functionality in reducing heart rate in a pilot study (n=2, male, aged ??) followed by a full study (n=18, 10 male, 8 female, ages ??) both with  participants not diagnosed with anxiety for the purposes of this preliminary robot system validation. The studies were reviewed by the school's ethics committee, and were conducted with parental consent and child agreement.
%The participants had their heart rate monitored in two conditions: 1) when interacting with the robot; and 2) when relaxing. Both studies revealed a significant decrease (p<.01) in heart rate during robot interaction compared to the relaxation condition. The extended study showed an average decrease of the children's' heart rate of 3.56 bpm (SD = 7.32 bpm).

We evaluated the robot in two user studies with children aged 7 to 8 and not diagnosed with anxiety.  The first was a pilot study that involved two participants and the second was the main study with 18 participants.  Both studies compared the participants' heart rates while playing with the robot and while relaxing because heart rate is a reliable physiological indicator of stress and relaxation \cite{Pittella2018WearableHR, DAlvia2020HeartRM, Sadeghi2019AnalyzingHR, Regula2014StudyOH}. The results  across both studies indicate that pocket-sized therapeutic robots have the potential to induce relaxation in children and promote this robot's potential as a calming tool.

\section{Related Work}
\subsection{Anxiety disorder}
Anxiety disorder manifests differently with each child, but often includes fear of going to school, social isolation, and a generally decreased quality of life \cite{VanAmeringen2003TheIO, Shields2004SocialAD, Ollendick2002TheDP, Association2022DiagnosticAS, deLijster2018SocialAA}. Those suffering from social anxiety disorders often also have a heightened tendency to promote themselves as low-functioning individuals \cite{Stein2000DisabilityAQ}, and have a high risk of developing post-childhood depression \cite{Beesdo2007IncidenceOS}. Although these are severely impacting consequences, only a fraction of those with self-observed symptoms seek professional help statistically after 15-20 years suffering from symptoms \cite{Rasouli2023CoDesignOA}.
Traditional cognitive behavioral therapy (CBT) in early adolescence is beneficial in preventing some of the post-childhood symptoms\cite{Leigh2015CognitiveTF,Morgan1999DoesRS} but may not be accessible or effective for all children.

Previous research in anxiety management and self regulation covers a vast range of varied techniques. These techniques include meditation as used with nursing students \cite{Ratanasiripong2015StressAA}, mantra-based meditation \cite{lvarezPrez2022EffectivenessOMq}, self-talk, deep breathing, progressive muscle relaxation, and imagery \cite{Hernowo2021TheEO}. Exercise has also been explored as an effective addition to CBT \cite{Servant2019NonpharmacologicalTF}. It was also found that an effort to control impulsive responses was significantly associated with lower anxiety scores \cite{Silva2023SocialAA}.
Earlier research also suggests that the coping strategies promoted as a part of CBT, many were proven to have an equivalent impact \cite{Lichstein1988ClinicalRS}. 

As a complement to existing therapeutic techniques, we investigate the use of a pocket robot as a way of promoting a relaxed state through an attention diverting interaction that requires little training to have an effect.

\subsection{Therapeutic robots}
Socially assistive robots (SAR)  \cite{seifer_mataric2005} and other types of therapeutic robots have been used both to offer comfort for children (with shapes as the pillow shaped TACO robot) \cite{OBrien2021ExploringTD}) and to provide meaningful pet-inspired interactions to lessen pain (such as with Paro the seal robot \cite{Pu2019HowPW}). SAR robots have also been used to reduce pain in children by offering empathetic verbal responses \cite{trostmataric2020}. A review paper from 2019 of using socially assistive robotics to aid children in distress and pain from the same authors suggested that using robots in such scenarios yielded a high interest and found that anxiety was reduced in the one of the included studies to a significant extent \cite{trostmataric2019}.

SAR have also been used for interventions for those suffering from social anxiety, with results outlining both risks and benefits \cite{Rasouli2022PotentialAO}. Robots may be best employed when used to compliment validated therapy. Examples of robots used in a coaching role have in a similar direction been attempted to address fear of public speaking \cite{Rasouli2023CoDesignOA}. 
%While not a physical robot, Salza et al. 2020 investigated the impact of using therapist-assisted computerized cognitive behavioral therapy (TacCBT), while Etzelmueller et al. 2020 investigated Internet-based intervention methods. They both found that the alternatives to person-to-person CBT had an impact on the anxiety patients and could be considered effective\cite{Salza2020CognitiveBT, Etzelmueller2020EffectsOI}. 
Nomura et al. 2019 found that users with high-anxiety tended to feel less anxious when they knew that they would be interacting with a robot 
\cite{Nomura2019DoPW}. 
%An important aspect of succesfull CBT is to motivate the patients because lasting results are often gained when the patients are personally invested in forming the change \cite{Ryan2011MotivationAA}.
%Although not measured in this paper, the robot we present in this paper and the tactile interaction has been developed with the intention of also increasing motivation for the children who interact with it. 

Inspired by past work on therapeutic robots, and aiming for a low-cost and broadly accessible approach for the anxiety context, we designed AffectaPocket, described next.
\section{Technical Overview}
AffectaPocket (shown in Figure \ref{affecta_annotated}) is a minimalistic robot, consisting of a 3D printed plastic body, tactile button sensors, a controller, a vibration actuator, and a small screen. We call AffectaPocket a robot and refer to it as such when working with children, in order to promote interest and engagement.

The robot's physical design features a 3D-printed outer shell, an ESP32 as its main processor. It is equipped with four tactile buttons:  two on the front and two activated by grasping its sides.  The 170 x 270 pixel screen serves multiple purposes. The upper displays simple eyes for engagement while the lower portion displays functional messages (e.g., charging status) and visual cues during the game's tutorial mode. 

The intended therapeutic functionality of AffectaPocket is stimulated through tactile interactions. Users interact with it by grasping its sides, which initiates a rhythm-matching game when held for three seconds. The game distracts the child by challenging them to replicate a randomly generated 3-note vibration pattern by grasping the robot's sides in a similar rhythmic pattern. 

A tutorial mode, where users can view the rhythmic pattern with visual cues (as seen in Figure \ref{visual_cues}) displayed with stars on the robot's screen aids children in creating a mental model of the game. To maintain discretion, the robot is silent, providing a source of support that can be kept concealed and private and used as needed.

\section{Pilot Study}
When designing assistive devices for vulnerable user populations, ensuring robust functionality is even more important than usual. Initially testing with non-diagnosed participants can reveal potential design flaws and inform necessary refinements. With this in mind, we tested the robot with non-diagnosed children aged 5-7 to gain valuable insights into the robot's functionality, allowing us to identify issues before potentially proceeding with our plans for evaluations with children experiencing anxiety disorders.

\subsection{Method} 
\begin{figure}[h]
\centering \includegraphics[width=0.50\textwidth]{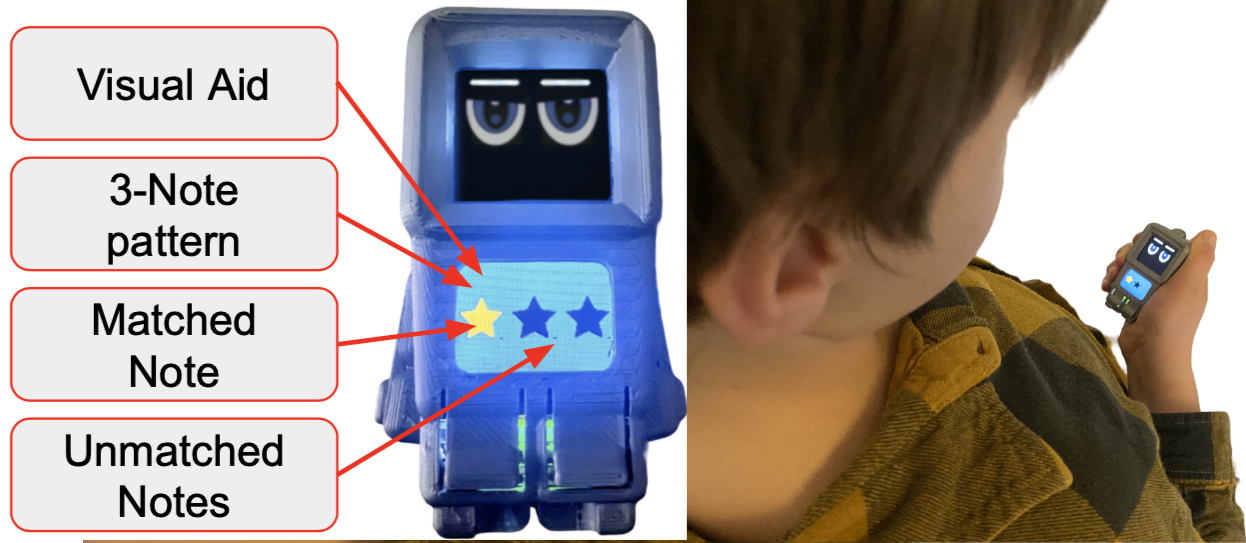} \caption{LEFT: The visual cues used to teach the rules of the rhythm-matching game. The stars appear one by one following the generated rhythmic pattern. The participants attempt to match the rhythm by grasping the side buttons. Matched rhythm steps appear as yellow stars, while unmatched steps appear as black stars. RIGHT: A child interacts with the robot in the tutorial mode with visual cues. The visual cues were used solely in the tutorial and not during the experiment.}
\label{visual_cues}
\end{figure}

A pilot study investigated the impact of the robot's interactions on participants' heart rate. Two children selected for their age-appropriateness and potential responsiveness to the intervention participated in a within-subjects design with randomized condition order. Over 14 consecutive days, both participants were exposed to the following conditions, administered at their home once daily at the same time (between 3pm and 4pm), for one minute, in random order with a 20 minute timed break between conditions:
\begin{itemize}
    \item \textbf{With-Game}: Participants were asked to play the tactile game with the robot.
    \item \textbf{No-game (Non-use)}: Participants were asked to hold the robot with the tactile game disabled.
\end{itemize}
We opted for the two conditions both using the robot to gain insight into the specific impact of the robot's tactile communication functionality. The dependent variable of the study was heart rate, measured in beats per minute (bpm). While heart rate variability (HRV) may be a stronger stress indicator, precise measurement requires longer exposure periods and extended baseline measurements so HRV was not applicable for this work \cite{Michels2013ChildrensHR, Khanade2017EfficacyOU, Nada2009HeartRV}. We chose heart rate as a measure for evaluating the immediate impact of different conditions. It was continuously recorded using a Polar heart rate tracker. Participants wore the tracker on their dominant arm while standing up. It is important to state that the visual cues (as seen in Figure \ref{visual_cues}) on the robot's torso screen were solely used during the initial presentation of the robot to the child. The study compares two conditions: using the robot's tactile game and holding the robot with no game. In neither condition is any visual input communicated to the child during the test. The robot is concealed in the pocket and is therefore out of sight for the child. The only difference between the two conditions is the presence of tactile vibrations, isolated from other communication modalities.

\subsection{Experiment design}
On the first day of the 14-day pilot study, the two participants were introduced to the game through the tutorial using the screen (as displayed in Figure \ref{visual_cues}).  Then the data collection phase began.  Each day, the process was as follows: 

\begin{itemize}
    \item The participant was fitted with the hear rate tracking band.
    \item The participants was asked to stand on a spot marked on the ground with red tape. 
    \item Heart rate tracking was initiated.
    \item The participant then completed the following in random order:\\ 1. Game condition: initiate the tactile interaction game by grasping the robot's side buttons for three seconds (With-game condition).\\ 2. No-game condition: stand idle for one minute.
    \item A timed 20 minute break was held in between the two conditions.
\end{itemize}

The experiment was repeated every day for fourteen days, alternating the order of the conditions.

\subsection{Results}

\begin{figure}[h]
\centering \includegraphics[width=0.50\textwidth]{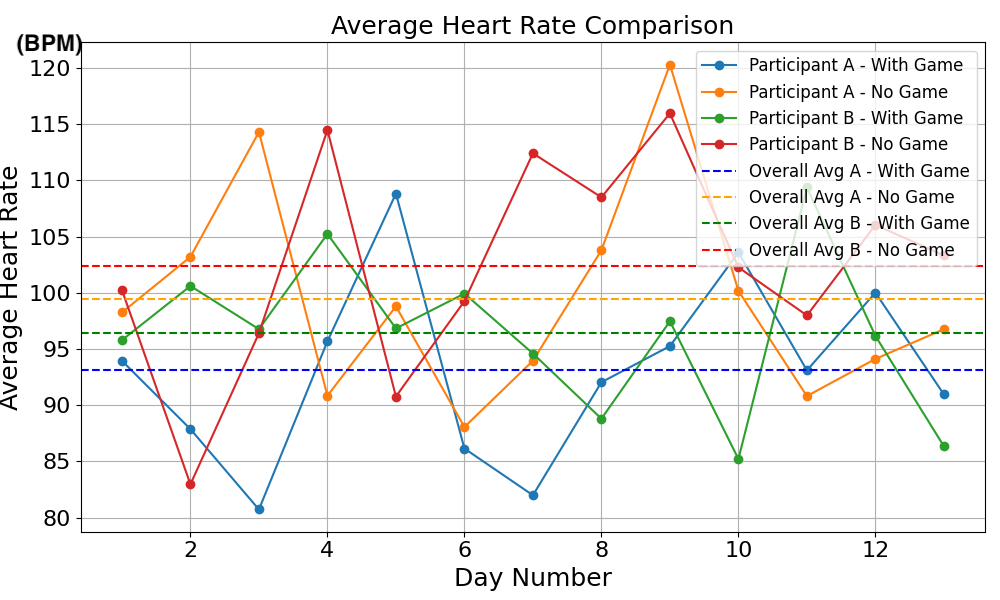} \caption{Pilot Study: The series represent "Participant A - with game," showing the average heart rates per session using the tactile game, while "Participant A - no game" displays the average heart rates without using the tactile game. The same data is shown for Participant B. The horizontal lines display the average heart rates for with-game and no-game across 14 days of measurements (of approximately 930 individual measurements). For both participants, the average values decrease when participants interact with the robot.}
\label{pilot-results}
\end{figure}

Figure \ref{pilot-results} shows the average heart rate (in beats per minute BPM) by test-condition over 14 days.  Consistently lower average heart rates were found in the "with-game" condition for both participants. Paired samples t-tests confirmed a statistically significant difference in heart rate between conditions for participants (p$<$.01). 
Parting the gathered data into two phases (of 7 days) the average decrease in heart rate between the two conditions grows from 4,22bpm in the first half to 8,15bpm in the second half indicating that the effect of interacting with the robot increases over time. However, more participants over a similar length of time would be needed to confirm this tendency.

\section{Main Study}
 Building on the pilot study, we conducted a main study with a larger sample size. We recruited 18 first-grade school children (ages 7-8) from a local Danish elementary school.  Informed consent was obtained from the parents of the participants, and verbal assent was given by each child before commencing the experiment. Similar to the pilot study, we employed the same within-subjects design with randomized condition order and each participant performed the steps outlined in the pilot study, except that the study was conducted at the school, in an empty classroom, between 9am and 1pm, over three consecutive days, and the exposure time of each condition was doubled to two minutes.

\subsection{Results}
\begin{figure}[h] \centering \includegraphics[width=0.50\textwidth]{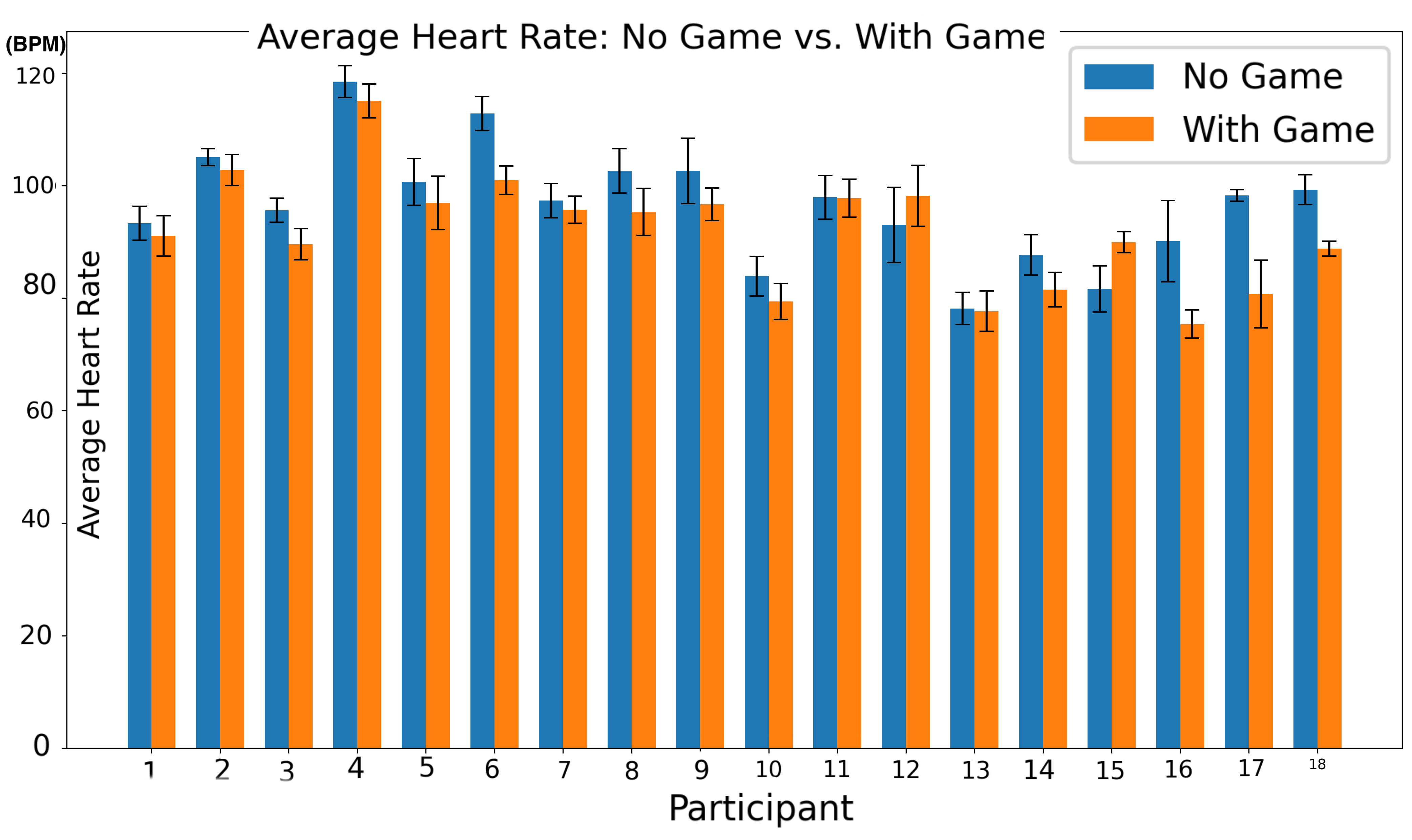} \caption{The average difference in heart rate for each of the conditions per participant; blue bars indicate the no-game condition and orange bars the with-game condition.}
\label{main_study}
\end{figure}

Figure \ref{main_study} displays the average heart rate for each participant during the two-minute "with-game" and "no-game" conditions. To analyze the difference between conditions, we calculated the within-participant heart rate change for each participant.  A Shapiro-Wilk normality test confirmed that these difference scores were normally distributed (W' = 0.0008). The average heart rate difference was 3.56 bpm, indicating a decrease during the "with-game" condition, with a standard deviation of 7.32 bpm.

A paired-samples t-test on the original "with-game" and "no-game" measurements revealed a highly significant difference (p$<$.01), supporting the observation of lower heart rates during active play interactions with the robot.

We calculated the effect size using Cohen's d (= 0.324836) confirming that there was a noticeable but medium sized difference between the conditions.

\section{Discussion}

\subsection{Heart rate decrease}

The positive Cohen's d (=0.324836) confirms a lower heart rate during the "with-game" condition compared to the "no-game" condition, with the average decrease of 3.56 bpm. Even a small to medium effect on heart rate carries physiological implications and a consistent decrease suggests potential benefits and warrants further investigation.

While the average decrease of 3.56 bpm is modest, it should be considered in the broader context of heart rate and relaxation research. 
Previous studies on controlled breathing exercises report similar heart rate reductions over extended periods (e.g., 20 minutes) \cite{Garg2023EffectOB, Shao2024TheEO, turankar2013}. Our findings demonstrate that a comparable decrease can be achieved through a brief, two-minute, game-based intervention without prior training. This underscores the potential of robots like AffectaPocket, offering an accessible option to supplement other methods, especially for individuals who may find other relaxation techniques challenging due to their required training and practice. One of the tested conditions in the experiments involved a monotonous task with somewhat stringent requirements to stand idle. This can be especially challenging for children to accomplish, and depending on how active the children are during this condition, it can potentially have an effect on their heart rate. We did not observe any overactive behaviors from the children during the study, but we acknowledge that future studies could benefit from including tasks for the children to prevent heart rate-increasing activities during the no-game conditions.
 
\subsection{Scaling the effect}

We found that the participants' heart rate decreased more in the longer-term (14-day) pilot study, with an average difference in bpm for the two conditions each day of 4.38 bpm than in the main (3-day) study, with an average of 3.56 bpm. This potential trend indicates that the robot's relaxing effect might increase as users become more accustomed to the interaction, but cannot be concluded due to the small sample size.

Children's initial excitement about the robot could have contributed to elevated heart rates, which naturally decreased as the novelty wore off.  We observed similar excitement during the extended study. However, even with this initial excitement effect, a significant difference in the average heart rate was still evident for the observations during the with-robot condition compared to the no-robot condition. This reinforces the potential benefits of the intervention.

While practical constraints prevented a longer-term study with the larger participant group,  future research should explore whether the effect size indeed  increases over time, over a longer-term study with a larger participant cohort.

\subsection{Heart rate measures}
Heart rate data is inherently sensitive to contextual factors, and the magnitude of heart rate change can differ between individuals based on their physical shape, emotional state, and recent activity levels. While our randomized within-subjects design aimed to control for some of this variation, individual differences likely still influenced the observed results.

The result remains significant across all participants. However,  interacting with the robot likely affected some participants more than others. The standard deviation of 7.32 bpm in the difference between "no-game" and "with-game" conditions demonstrates that some participants experienced a substantial impact from using the robot, while others were less affected.  Further research should explore the internal and external factors that might contribute to this variability in responses to better understand and optimize the robot's therapeutic effects.
 
Overall, these early results suggest that our novel therapeutic pocket robot design holds promise for inducing calmness. While further research is needed, our findings indicate that pocket-sized therapeutic robots present a viable option for promoting relaxation. These results support the need for future studies specifically investigating the efficacy of this technology with vulnerable populations experiencing anxiety disorders.
\section{Conclusion}

This paper investigated the physical impact of interacting with a tactile, pocket-sized therapeutic robot designed to promote a relaxed and focused state. We presented a small therapeutic robot with functionality to introduce an increased relaxed state by diverting the attention of its users. This intervention utilizes a tactile rhythm game as a way to interact with the robot.

The robot was initially tested in a 14-day pilot study with two participants. This study focused on heart rate as a measure of relaxation to evaluate the robot's calming potential. We observed a significant decrease in heart rate when the participants interacted with the robot. This finding was confirmed and strengthened in an expanded study with 18 healthy children. These results suggest considerable potential for using pocket-sized therapeutic robots as a relaxation tool and warrant further research with children diagnosed with anxiety disorders.

\balance

\bibliography{bibliography}
\bibliographystyle{IEEEtran}

\end{document}